\documentclass[conference]{IEEEtran}
\usepackage{times}

\usepackage[numbers]{natbib}
\usepackage{multicol}
\usepackage[bookmarks=true]{hyperref}
\usepackage{xcolor}
\newcommand{\pkg}[1]{\textsc{#1}}

\begin{document}

\title{Constrained Dynamics Simulation: More With Less}

\author{Ajay Suresha Sathya
\thanks{\textsuperscript{1}~Inria - Département d’Informatique de l’École normale supérieure, PSL Research University. {\tt\small \{ajay.sathya\}@inria.fr}}}

\maketitle

\IEEEpeerreviewmaketitle









\section{Introduction}
Efficient robot dynamics simulation is a fundamental problem key for robot control, identification, design and analysis. Efficient simulation is a crucial enabler of model predictive control (MPC)~\cite{rawlings2017model} and learning-based control~\cite{lee2020learning}, arguably two of the most successful and currently the most actively researched advanced robot control techniques. It can improve the optimality and safety properties of MPC by enabling a longer prediction horizon~\cite{rawlings2017model}, enabling highly dynamic robot behaviour~\footnote{https://www.youtube.com/watch?v=tF4DML7FIWk} close to system limits. Efficient simulation can shorten training times of learning-based controllers, such as a reinforcement learning (RL)~\cite{sutton2018reinforcement,bertsekas2019reinforcement} based controller, where the learning phase often relies heavily on simulation~\cite{lee2020learning} due to the prohibitive cost and safety concerns involved with training on real robots. Faster and physically accurate simulators can democratize robotics research by potentially not requiring dedicated expensive resources like high-end GPUs and physical robots for training and validating controllers. Accurate simulation is also key to task and motion planning (TAMP)~\cite{toussaintdifferentiable, garrett2021integrated} that will be essential for intelligent robots executing long-horizon tasks robustly. Finally, simulation profoundly impacts several adjacent fields like biomechanics~\cite{delp2007opensim} and computer graphics~\cite{coumans2021}.

\noindent \textbf{Current simulator efficiency is insufficient.} Simulation is often the bottleneck, particularly in MPC, where evaluating dynamics and derivatives can take up to 80\% of the total computation cost~\cite{astudillo2021mixed}. Consquently, achieving whole-body MPC for high degree of freedom (DoF) systems like humanoids remains an open problem. Moreover, industrial applications of solving minimum-time optimal control problems with full dynamics models, to maximize economic objectives like bin picks-per-hour~\cite{sathya2022tasho}, is also limited perhaps due to the long solution times~\cite{howell2019altro} offsetting the performance gains. Moreover, simulation inefficiencies force roboticists to ignore submechanisms like gears~\cite{jain1990recursive} or joint flexibility~\cite{albu2007unified}, which causes significant simulation errors~\cite{chignoli2023recursive}. Inspired by the large language models' (LLMs)~\cite{achiam2023gpt,touvron2023llama} success, early works have trained large multimodal models targetting robotics~\cite{driess2023palm,brohan2023rt}. Simulation is attractive~\cite{eppner2021acronym} for generating sufficient amount of diverse data for training these data-hungry models. Considering the scale of the typical datasets~\cite{brohan2023rt}, reducing the carbon footprint of generating training data and evaluating trained models in simulation, assumes key importance. 

\noindent \textbf{Drawbacks of existing simulators.} 
Solving the inner equality constrained dynamics problem~\cite{featherstone2014rigid} is often the most computationally expensive aspect of simulating contact dynamics. To solve these inner problems, most existing simulators (\pkg{RBDL}~\cite{felis2017rbdl}, \pkg{Raisim}~\cite{hwangbo2018per}, \pkg{MuJoCo}~\cite{todorov2012mujoco,todorov2014convex}, \pkg{Pinocchio}~\cite{carpentier2019pinocchio}, \pkg{Drake}~\cite{drake}, \pkg{Dart}~\cite{lee2018dart}, \pkg{Bullet}~\cite{coumans2021} and PhysX~\footnote{https://developer.nvidia.com/physx-sdk} to name a few) use Featherstone's branching-induced sparsity-exploiting LTL algorithm~\cite{featherstone2005efficient,featherstone2010exploiting}, which however has an expensive computational complexity of $O(nd^2 + m^2d + dm^2 + m^3)$, where $n$, $d$ and $m$ are the robot's DoF, kinematic tree's depth, and constraint dimensionality respectively. This worst-case cubic complexity renders the LTL algorithm inefficient for high DoF robots.  Other recent simulators like BRAX~\cite{freeman2021brax} adopt a computationally even-worse approach of Gauss-Seidel iterations to solve the KKT system arising from the inefficient~\cite{featherstone2014rigid} maximal coordinate formulation of Baraff~\cite{baraff1996linear}. But they mitigate their algorithmic inefficiency to an extent by exploiting the compute power of GPUs or TPUs. 
However, there exist more efficient recursive algorithms~\cite{popov1978manipuljacionnyje,vereshchagin1989modeling,bae1987recursive,otter1987algorithm} in dynamics literature that have been forgotten or ignored that have linear complexity in $n$. 

These low-complexity recursive algorithms offer a unique and untapped opportunity to accelerate existing simulators. However, they require a revival and possibly an improvement. The low-complexity algorithms require non-trivial extensions to efficiently support closed-loop mechanisms. An efficient {\tt C++} implementation is also essential to fully exploit an efficient algorithm. A simulator for robotics should be physically realistic and not ignore complementarity constraints~\cite{todorov2014convex} or linearize the nonlinear complementarity problem~\cite{coumans2021} (NCP) associated with the frictional contact problem, as this realism can potentially require less domain randomization or control feedback to deal with simulator error. Finally, the simulator should also be differentiable with smoothing~\cite{suh2022differentiable} without sacrificing efficiency to enable trajectory optimization and learning for contact-rich tasks. None of the existing simulators satisfy all the requirements listed in this paragraph and building one is my current research problem. 

\section{Current Research}

\noindent \textbf{Linear complexity constrained dynamics algorithms:}
I derived a family of efficient constrained dynamics algorithms (CDAs)~\cite{sathya_constrained_dynamics} for kinematic trees by solving an equivalent discrete-time linear quadratic regulator (LQR) problem~\cite{rawlings2017model}, arising from Gauss' principle of least constraint~\cite{gauss1829neues,udwadia1996,bruyninckx2000gauss}. I first revisited and revived the efficient, but largely unknown, Popov-Vereshchagin (PV) CDA~\cite{popov1978manipuljacionnyje,vereshchagin1989modeling} from 1970s, with a complexity of $O(n + m^2d + m^3)$. Since the original paper~\cite{vereshchagin1989modeling} has no derivation, I provided an expository derivation in~\cite{sathya_constrained_dynamics} by adapting the textbook dynamic programming~\cite{bellman1966dynamic} approach to solve the corresponding LQR problem~\cite{rawlings2017model}. I further extended the PV algorithm to floating-base trees, with constraints permitted on any link, and updated it with modern efficiency techniques discovered over the years like adding uniform gravity field~\cite{brandl1986very}, using local frames~\cite{brandl1986very} and using DH nodes~\cite{mcmillan1995efficient}. 
Then, by exploring LQR solver variants, I proposed two new CDAs~\cite{sathya_constrained_dynamics} with asymptotically optimal $O(n+m)$ complexity, namely PV-soft and PV-early, derived using \pkg{MuJoCo}-style relaxed constraints and aggressive early elimination of Lagrange multipliers respectively. My numerical results indicate significant simulation speed-up of 2x for high dimensional
robots likquadrupeds and humanoids.

The LQR/optimization connection in my derivation makes CDAs accessible to researchers with control and optimization background, which includes many roboticists due to the popularity of OCP solvers~\cite{tassa2014control,farshidian2017efficient,mastalli2020crocoddyl}, who currently use CDAs as a black-box. This connection facilitated my follow-up work~\cite{sathya2024constrained} on proximal~\cite{parikh2014proximal} dynamics algorithms three original dynamics algorithms, the most efficient among which is constrainedABA with $O(n+m)$ complexity. These proximal algorithms relax the constraint linear independence assumption in the PV solvers, are numerically robust to singular cases, automatically return least-squares solution for even infeasible motion constraints~\cite{chiche2016augmented,guler1991convergence}. They generalize existing algorithms, namely PV-soft and the CDAs in \pkg{MuJoCo} and \pkg{Drake} allowing trading-off compliance and rigidity in contact through proximal iterations. ConstrainedABA is particularly simple and easily implemented requiring only a few additional lines of code compared to Featherstone's articulated-body algorithm~\cite{featherstone1983calculation,featherstone2014rigid}, faciliating its implementation in existing simulators.



\noindent \textbf{Lowest complexity Delassus matrix algorithms:}
The Delassus matrix~\cite{delassus1917memoire,duriez2005realistic} or the inverse operational space inertia matrix (OSIM)~\cite{khatib1987unified} is key to efficiently differentiating~\cite{carpentier2019pinocchio,nganga2023accelerating} CDAs and solving frictional contact problems~\cite{duriez2005realistic,raisim,lidec2023contact}. I discovered that the PV algorithm computes the Delassus matrix as an intermediate quantity, thereby providing a new Delassus matrix algorithm (PV-OSIM), with a computational complexity of $O(n+m^2d+m^2)$. I further accelerated PV-OSIM for floating-base robots with branching at the base using the matrix inversion lemma~\cite{sherman1950adjustment}. PV-OSIM was found to be significantly faster for most realistic robots (upto 2x for humanoids) than the existing state-of-the-art recursive algorithms, KJR~\cite{kreutz1992recursive} and EFPA~\cite{wensing2012reduced}, which have a complexity of $O(n+m^2d+m^2)$ and $O(n+md+m^2)$. In a follow-up work~\cite{sathya_low_order_delassus}, I exploited the compositionality of the extended force and
motion propagators~\cite{lilly1989efficient,chang2001efficient} in PV-OSIM computation to obtain the PV-OSIMr algorithm,  with an optimal complexity of $O(n+m^2)$. However, computing constraint forces requires factorizing the Delassus matrix, which incurs an additional $O(m^3)$ operations for all the above algorithms. The combination of proximal algorithms and the matrix inversion lemma in my recent work~\cite{sathya2024constrained}, yields a new algorithm cABA-OSIM, that can compute even the damped Delassus inverse matrix in just $O(n+m^2)$ operations. cABA-OSIM was found to be over $3x$ faster for humanoids than the widely-used Featherstone's LTL-OSIM algorithm~\cite{featherstone2010exploiting,carpentier2019pinocchio}. 

\noindent \textbf{Implementation:} All the low-complexity dynamics algorithms mentioned above have been recently implemented in {\tt C++} within the high-quality open-source dynamics library \pkg{Pinocchio}~\cite{carpentier2019pinocchio} and will shortly be released to the community. \pkg{Pinocchio} was chosen because its efficient implementation alone provides several times speed-up compared to other simulators~\cite{coumans2021,lee2018dart} also written in {\tt C++}.


\section{Future research}

\noindent \textbf{Short term:}
There exist several pressing extensions and clear short-term future research directions. Mechanisms with \textbf{internal closed-loops}~\cite{jain2010robot} are inadequately addressed in most existing simulators and recent robot designs increasingly use internal loops~\cite{kangaroo,digit} for mechanical reasons. We have made initial progress on extending the proximal dynamics algorithms~\cite{sathya2024constrained} to closed-loop mechanisms. Next, these low-complexity algorithms will be used to solve \textbf{inequality constraints} and \textbf{frictional contact problems}, thereby delivering a fully-fledged efficient simulator using our low-complexity algorithms. The lessons learned in~\cite{lidec2023contact} will be used to ensure the physical realism of the frictional contact simulator.


Making our low-complexity simulator \textbf{differentiable} is the natural next step. I expect the derivative computation to be significantly accelerated due to my cABA-OSIM algorithm and through a {\tt C++} implementation by adopting the implicit function approach~\cite{carpentier2018analytical}. Note that using proximal algorithms makes differentiation through our proximal CDAs well-defined (over finite number of iterations) even for singular/infeasible cases due to the differentiability of the proximal operator.  To obtain informative gradients despite the non-smoothness inherent in frictional contact problems, I plan to explore and support the randomized smoothing methods~\cite{suh2022bundled,suh2022differentiable,le2024leveraging} with our low-complexity algorithms.

\noindent \textbf{Medium and long term:} In the medium and long-term, my work will explore applications and opportunities enabled by the fast and differentiable simulator. Armed with the cumulative speed-ups we expect due to algorithmic improvements as well as efficient implementation, a medium-term direction is to push towards \textbf{whole-body MPC} for humanoid-sized robots. For long term, leveraging the gradient information from the simulator to reduce the \textbf{sample complexity of RL} or imitation learning, where the current models and learning methods remain data-hungry~\cite{jenelten2024dtc}, is another potentially interesting direction. This can be critical for manipulation tasks where there is arguably a higher diversity of challenges compared to the locomotion tasks, making data generation challenging. Finally, another exciting longer term direction is enabling \textbf{long-horizon planning}~\cite{mokhtari2021safe} for set of tasks requiring fast and dynamic tool use~\cite{toussaintdifferentiable} using physically realistic simulators.

Finally, through efficient algorithms and implementations, I hope to democratize robotics research by enabling prototyping and deployment of advanced control techniques like MPC and RL even on resource-constrained computational platforms.

\bibliographystyle{plainnat}
\bibliography{references}

\begin{thebibliography}{71}
\providecommand{\natexlab}[1]{#1}
\providecommand{\url}[1]{\texttt{#1}}
\expandafter\ifx\csname urlstyle\endcsname\relax
  \providecommand{\doi}[1]{doi: #1}\else
  \providecommand{\doi}{doi: \begingroup \urlstyle{rm}\Url}\fi

\bibitem[dig()]{digit}
Digit robot.
\newblock URL \url{https://agilityrobotics.com/robots}.

\bibitem[kan()]{kangaroo}
Kangaroo robot.
\newblock URL \url{https://pal-robotics.com/robots/kangaroo/}.

\bibitem[Achiam et~al.(2023)Achiam, Adler, Agarwal, Ahmad, Akkaya, Aleman,
  Almeida, Altenschmidt, Altman, Anadkat, et~al.]{achiam2023gpt}
Josh Achiam, Steven Adler, Sandhini Agarwal, Lama Ahmad, Ilge Akkaya,
  Florencia~Leoni Aleman, Diogo Almeida, Janko Altenschmidt, Sam Altman,
  Shyamal Anadkat, et~al.
\newblock Gpt-4 technical report.
\newblock \emph{arXiv preprint arXiv:2303.08774}, 2023.

\bibitem[Albu-Sch{\"a}ffer et~al.(2007)Albu-Sch{\"a}ffer, Ott, and
  Hirzinger]{albu2007unified}
Alin Albu-Sch{\"a}ffer, Christian Ott, and Gerd Hirzinger.
\newblock A unified passivity-based control framework for position, torque and
  impedance control of flexible joint robots.
\newblock \emph{The international journal of robotics research}, 26\penalty0
  (1):\penalty0 23--39, 2007.

\bibitem[Astudillo et~al.(2021)Astudillo, Carpentier, Gillis, Pipeleers, and
  Swevers]{astudillo2021mixed}
Alejandro Astudillo, Justin Carpentier, Joris Gillis, Goele Pipeleers, and Jan
  Swevers.
\newblock Mixed use of analytical derivatives and algorithmic differentiation
  for nmpc of robot manipulators.
\newblock \emph{IFAC-PapersOnLine}, 54\penalty0 (20):\penalty0 78--83, 2021.

\bibitem[Bae and Haug(1987)]{bae1987recursive}
Dae-Sung Bae and Edward~J Haug.
\newblock A recursive formulation for constrained mechanical system dynamics:
  Part ii. closed loop systems.
\newblock \emph{Journal of Structural Mechanics}, 15\penalty0 (4):\penalty0
  481--506, 1987.

\bibitem[Baraff(1996)]{baraff1996linear}
David Baraff.
\newblock Linear-time dynamics using lagrange multipliers.
\newblock In \emph{Proceedings of the 23rd annual conference on Computer
  graphics and interactive techniques}, pages 137--146, 1996.

\bibitem[Bellman(1966)]{bellman1966dynamic}
Richard Bellman.
\newblock Dynamic programming.
\newblock \emph{Science}, 153\penalty0 (3731):\penalty0 34--37, 1966.

\bibitem[Bertsekas(2019)]{bertsekas2019reinforcement}
Dimitri Bertsekas.
\newblock \emph{Reinforcement learning and optimal control}.
\newblock Athena Scientific, 2019.

\bibitem[Brandl et~al.(1986)Brandl, Johanni, and Otter]{brandl1986very}
Helmut Brandl, Rainer Johanni, and Martin Otter.
\newblock A very efficient algorithm for the simulation of robots and similar
  multibody systems without inversion of the mass matrix.
\newblock \emph{IFAC Proceedings Volumes}, 19\penalty0 (14):\penalty0 95--100,
  1986.

\bibitem[Brohan et~al.(2023)Brohan, Brown, Carbajal, Chebotar, Chen,
  Choromanski, Ding, Driess, Dubey, Finn, et~al.]{brohan2023rt}
Anthony Brohan, Noah Brown, Justice Carbajal, Yevgen Chebotar, Xi~Chen,
  Krzysztof Choromanski, Tianli Ding, Danny Driess, Avinava Dubey, Chelsea
  Finn, et~al.
\newblock Rt-2: Vision-language-action models transfer web knowledge to robotic
  control.
\newblock \emph{arXiv preprint arXiv:2307.15818}, 2023.

\bibitem[Bruyninckx and Khatib(2000)]{bruyninckx2000gauss}
Herman Bruyninckx and Oussama Khatib.
\newblock Gauss' principle and the dynamics of redundant and constrained
  manipulators.
\newblock In \emph{Proc. IEEE Int. Conf. Robot. Autom.}, volume~3, pages
  2563--2568. IEEE, 2000.

\bibitem[Carpentier and Mansard(2018)]{carpentier2018analytical}
Justin Carpentier and Nicolas Mansard.
\newblock Analytical derivatives of rigid body dynamics algorithms.
\newblock In \emph{Proc. Robot., Sci. Syst.}, 2018.

\bibitem[Carpentier et~al.(2019)Carpentier, Saurel, Buondonno, Mirabel,
  Lamiraux, Stasse, and Mansard]{carpentier2019pinocchio}
Justin Carpentier, Guilhem Saurel, Gabriele Buondonno, Joseph Mirabel, Florent
  Lamiraux, Olivier Stasse, and Nicolas Mansard.
\newblock The pinocchio c++ library: A fast and flexible implementation of
  rigid body dynamics algorithms and their analytical derivatives.
\newblock In \emph{2019 IEEE/SICE International Symposium on System Integration
  (SII)}, pages 614--619. IEEE, 2019.

\bibitem[Chang and Khatib(2001)]{chang2001efficient}
Kyong-Sok Chang and Oussama Khatib.
\newblock Efficient recursive algorithm for the operational space inertia
  matrix of branching mechanisms.
\newblock \emph{Advanced Robotics}, 14\penalty0 (8):\penalty0 703--715, 2001.

\bibitem[Chiche and Gilbert(2016)]{chiche2016augmented}
Alice Chiche and Jean~Charles Gilbert.
\newblock How the augmented lagrangian algorithm can deal with an infeasible
  convex quadratic optimization problem.
\newblock \emph{Journal of Convex Analysis}, 23\penalty0 (2), 2016.

\bibitem[Chignoli et~al.(2023)Chignoli, Adrian, Kim, and
  Wensing]{chignoli2023recursive}
Matthew Chignoli, Nicholas Adrian, Sangbae Kim, and Patrick~M Wensing.
\newblock Recursive rigid-body dynamics algorithms for systems with kinematic
  loops.
\newblock \emph{arXiv preprint arXiv:2311.13732}, 2023.

\bibitem[Coumans and Bai(2016--2021)]{coumans2021}
Erwin Coumans and Yunfei Bai.
\newblock Pybullet, a python module for physics simulation for games, robotics
  and machine learning.
\newblock \url{http://pybullet.org}, 2016--2021.

\bibitem[Delassus(1917)]{delassus1917memoire}
{\'E}tienne Delassus.
\newblock M{\'e}moire sur la th{\'e}orie des liaisons finies unilat{\'e}rales.
\newblock In \emph{Annales scientifiques de l'{\'E}cole normale
  sup{\'e}rieure}, volume~34, pages 95--179, 1917.

\bibitem[Delp et~al.(2007)Delp, Anderson, Arnold, Loan, Habib, John,
  Guendelman, and Thelen]{delp2007opensim}
Scott~L Delp, Frank~C Anderson, Allison~S Arnold, Peter Loan, Ayman Habib,
  Chand~T John, Eran Guendelman, and Darryl~G Thelen.
\newblock Opensim: open-source software to create and analyze dynamic
  simulations of movement.
\newblock \emph{IEEE transactions on biomedical engineering}, 54\penalty0
  (11):\penalty0 1940--1950, 2007.

\bibitem[Driess et~al.(2023)Driess, Xia, Sajjadi, Lynch, Chowdhery, Ichter,
  Wahid, Tompson, Vuong, Yu, et~al.]{driess2023palm}
Danny Driess, Fei Xia, Mehdi~SM Sajjadi, Corey Lynch, Aakanksha Chowdhery,
  Brian Ichter, Ayzaan Wahid, Jonathan Tompson, Quan Vuong, Tianhe Yu, et~al.
\newblock Palm-e: An embodied multimodal language model.
\newblock \emph{arXiv preprint arXiv:2303.03378}, 2023.

\bibitem[Duriez et~al.(2005)Duriez, Dubois, Kheddar, and
  Andriot]{duriez2005realistic}
Christian Duriez, Frederic Dubois, Abderrahmane Kheddar, and Claude Andriot.
\newblock Realistic haptic rendering of interacting deformable objects in
  virtual environments.
\newblock \emph{IEEE transactions on visualization and computer graphics},
  12\penalty0 (1):\penalty0 36--47, 2005.

\bibitem[Eppner et~al.(2021)Eppner, Mousavian, and Fox]{eppner2021acronym}
Clemens Eppner, Arsalan Mousavian, and Dieter Fox.
\newblock Acronym: A large-scale grasp dataset based on simulation.
\newblock In \emph{2021 IEEE International Conference on Robotics and
  Automation (ICRA)}, pages 6222--6227. IEEE, 2021.

\bibitem[Farshidian et~al.(2017)Farshidian, Neunert, Winkler, Rey, and
  Buchli]{farshidian2017efficient}
Farbod Farshidian, Michael Neunert, Alexander~W Winkler, Gonzalo Rey, and Jonas
  Buchli.
\newblock An efficient optimal planning and control framework for quadrupedal
  locomotion.
\newblock In \emph{2017 IEEE International Conference on Robotics and
  Automation (ICRA)}, pages 93--100. IEEE, 2017.

\bibitem[Featherstone(1983)]{featherstone1983calculation}
Roy Featherstone.
\newblock The calculation of robot dynamics using articulated-body inertias.
\newblock \emph{Int. J. Robot. Res.}, 2\penalty0 (1):\penalty0 13--30, 1983.

\bibitem[Featherstone(2005)]{featherstone2005efficient}
Roy Featherstone.
\newblock Efficient factorization of the joint-space inertia matrix for
  branched kinematic trees.
\newblock \emph{Int. J. Robot. Res.}, 24\penalty0 (6):\penalty0 487--500, 2005.

\bibitem[Featherstone(2010)]{featherstone2010exploiting}
Roy Featherstone.
\newblock Exploiting sparsity in operational-space dynamics.
\newblock \emph{Int. J. Robot. Res.}, 29\penalty0 (10):\penalty0 1353--1368,
  2010.

\bibitem[Featherstone(2014)]{featherstone2014rigid}
Roy Featherstone.
\newblock \emph{Rigid body dynamics algorithms}.
\newblock Springer, 2014.

\bibitem[Felis(2017)]{felis2017rbdl}
Martin~L Felis.
\newblock Rbdl: an efficient rigid-body dynamics library using recursive
  algorithms.
\newblock \emph{Autonomous Robots}, 41\penalty0 (2):\penalty0 495--511, 2017.

\bibitem[Freeman et~al.(2021)Freeman, Frey, Raichuk, Girgin, Mordatch, and
  Bachem]{freeman2021brax}
C~Daniel Freeman, Erik Frey, Anton Raichuk, Sertan Girgin, Igor Mordatch, and
  Olivier Bachem.
\newblock Brax--a differentiable physics engine for large scale rigid body
  simulation.
\newblock \emph{arXiv preprint arXiv:2106.13281}, 2021.

\bibitem[Garrett et~al.(2021)Garrett, Chitnis, Holladay, Kim, Silver,
  Kaelbling, and Lozano-P{\'e}rez]{garrett2021integrated}
Caelan~Reed Garrett, Rohan Chitnis, Rachel Holladay, Beomjoon Kim, Tom Silver,
  Leslie~Pack Kaelbling, and Tom{\'a}s Lozano-P{\'e}rez.
\newblock Integrated task and motion planning.
\newblock \emph{Annual review of control, robotics, and autonomous systems},
  4:\penalty0 265--293, 2021.

\bibitem[Gau{\ss}(1829)]{gauss1829neues}
Carl~Friedrich Gau{\ss}.
\newblock {\"U}ber ein neues allgemeines grundgesetz der mechanik.
\newblock 1829.

\bibitem[G{\"u}ler(1991)]{guler1991convergence}
Osman G{\"u}ler.
\newblock On the convergence of the proximal point algorithm for convex
  minimization.
\newblock \emph{SIAM journal on control and optimization}, 29\penalty0
  (2):\penalty0 403--419, 1991.

\bibitem[Howell et~al.(2019)Howell, Jackson, and Manchester]{howell2019altro}
Taylor~A Howell, Brian~E Jackson, and Zachary Manchester.
\newblock Altro: A fast solver for constrained trajectory optimization.
\newblock In \emph{2019 IEEE/RSJ International Conference on Intelligent Robots
  and Systems (IROS)}, pages 7674--7679. IEEE, 2019.

\bibitem[Hwangbo et~al.(2018{\natexlab{a}})Hwangbo, Lee, and
  Hutter]{hwangbo2018per}
Jemin Hwangbo, Joonho Lee, and Marco Hutter.
\newblock Per-contact iteration method for solving contact dynamics.
\newblock \emph{IEEE Robotics and Automation Letters}, 3\penalty0 (2):\penalty0
  895--902, 2018{\natexlab{a}}.

\bibitem[Hwangbo et~al.(2018{\natexlab{b}})Hwangbo, Lee, and Hutter]{raisim}
Jemin Hwangbo, Joonho Lee, and Marco Hutter.
\newblock Per-contact iteration method for solving contact dynamics.
\newblock \emph{{IEEE} Robot. Autom. Lett.}, 3\penalty0 (2):\penalty0 895--902,
  2018{\natexlab{b}}.
\newblock URL \url{www.raisim.com}.

\bibitem[Jain and Rodriguez(1990)]{jain1990recursive}
A~Jain and G~Rodriguez.
\newblock Recursive dynamics for geared robot manipulators.
\newblock In \emph{29th IEEE Conference on Decision and Control}, pages
  1983--1988. IEEE, 1990.

\bibitem[Jain(2010)]{jain2010robot}
Abhinandan Jain.
\newblock \emph{Robot and multibody dynamics: analysis and algorithms}.
\newblock Springer Science \& Business Media, 2010.

\bibitem[Jenelten et~al.(2024)Jenelten, He, Farshidian, and
  Hutter]{jenelten2024dtc}
Fabian Jenelten, Junzhe He, Farbod Farshidian, and Marco Hutter.
\newblock Dtc: Deep tracking control.
\newblock \emph{Science Robotics}, 9\penalty0 (86):\penalty0 eadh5401, 2024.

\bibitem[Khatib(1987)]{khatib1987unified}
Oussama Khatib.
\newblock A unified approach for motion and force control of robot
  manipulators: The operational-space formulation.
\newblock \emph{IEEE Journal on Robotics and Automation}, 3\penalty0
  (1):\penalty0 43--53, 1987.

\bibitem[Kreutz-Delgado et~al.(1992)Kreutz-Delgado, Jain, and
  Rodriguez]{kreutz1992recursive}
Kenneth Kreutz-Delgado, Abhinandan Jain, and Guillermo Rodriguez.
\newblock Recursive formulation of operational-space control.
\newblock \emph{Int. J. Robot. Res.}, 11\penalty0 (4):\penalty0 320--328, 1992.

\bibitem[Le~Lidec et~al.(2024)Le~Lidec, Schramm, Montaut, Schmid, Laptev, and
  Carpentier]{le2024leveraging}
Quentin Le~Lidec, Fabian Schramm, Louis Montaut, Cordelia Schmid, Ivan Laptev,
  and Justin Carpentier.
\newblock Leveraging randomized smoothing for optimal control of nonsmooth
  dynamical systems.
\newblock \emph{Nonlinear Analysis: Hybrid Systems}, 52:\penalty0 101468, 2024.

\bibitem[Lee et~al.(2018)Lee, X.~Grey, Ha, Kunz, Jain, Ye, S.~Srinivasa,
  Stilman, and Karen~Liu]{lee2018dart}
Jeongseok Lee, Michael X.~Grey, Sehoon Ha, Tobias Kunz, Sumit Jain, Yuting Ye,
  Siddhartha S.~Srinivasa, Mike Stilman, and C~Karen~Liu.
\newblock Dart: Dynamic animation and robotics toolkit.
\newblock \emph{The Journal of Open Source Software}, 3\penalty0 (22):\penalty0
  500, 2018.

\bibitem[Lee et~al.(2020)Lee, Hwangbo, Wellhausen, Koltun, and
  Hutter]{lee2020learning}
Joonho Lee, Jemin Hwangbo, Lorenz Wellhausen, Vladlen Koltun, and Marco Hutter.
\newblock Learning quadrupedal locomotion over challenging terrain.
\newblock \emph{Science robotics}, 5\penalty0 (47):\penalty0 eabc5986, 2020.

\bibitem[Lidec et~al.(2023)Lidec, Jallet, Montaut, Laptev, Schmid, and
  Carpentier]{lidec2023contact}
Quentin~Le Lidec, Wilson Jallet, Louis Montaut, Ivan Laptev, Cordelia Schmid,
  and Justin Carpentier.
\newblock {Contact Models in Robotics: a Comparative Analysis}.
\newblock \emph{arXiv preprint arXiv:2304.06372}, 2023.

\bibitem[Lilly(1989)]{lilly1989efficient}
Kathryn~Weed Lilly.
\newblock \emph{Efficient dynamic simulation of multiple chain robotic
  systems}.
\newblock The Ohio State University, 1989.

\bibitem[Mastalli et~al.(2020)Mastalli, Budhiraja, Merkt, Saurel, Hammoud,
  Naveau, Carpentier, Righetti, Vijayakumar, and
  Mansard]{mastalli2020crocoddyl}
Carlos Mastalli, Rohan Budhiraja, Wolfgang Merkt, Guilhem Saurel, Bilal
  Hammoud, Maximilien Naveau, Justin Carpentier, Ludovic Righetti, Sethu
  Vijayakumar, and Nicolas Mansard.
\newblock Crocoddyl: An efficient and versatile framework for multi-contact
  optimal control.
\newblock In \emph{2020 IEEE International Conference on Robotics and
  Automation (ICRA)}, pages 2536--2542. IEEE, 2020.

\bibitem[McMillan and Orin(1995)]{mcmillan1995efficient}
Scott McMillan and David~E Orin.
\newblock Efficient computation of articulated-body inertias using successive
  axial screws.
\newblock \emph{IEEE Transactions on Robotics and Automation}, 11\penalty0
  (4):\penalty0 606--611, 1995.

\bibitem[Mokhtari et~al.(2021)Mokhtari, Sathya, Tsiogkas, and
  Decr{\'e}]{mokhtari2021safe}
Vahid Mokhtari, Ajay~Suresha Sathya, Nikolaos Tsiogkas, and Wilm Decr{\'e}.
\newblock Safe-planner: A single-outcome replanner for computing strong cyclic
  policies in fully observable non-deterministic domains.
\newblock In \emph{2021 20th International Conference on Advanced Robotics
  (ICAR)}, pages 974--981. IEEE, 2021.

\bibitem[Nganga and Wensing(2023)]{nganga2023accelerating}
John~N Nganga and Patrick~M Wensing.
\newblock Accelerating hybrid systems differential dynamic programming.
\newblock \emph{ASME Letters in Dynamic Systems and Control}, 3\penalty0
  (1):\penalty0 011002, 2023.

\bibitem[Otter et~al.(1987)Otter, Brandl, and Johanni]{otter1987algorithm}
M~Otter, H~Brandl, and R~Johanni.
\newblock An algorithm for the simulation of multibody systems with kinematic
  loops.
\newblock In \emph{Proceedings of the 7th World Congress on Theory of Machines
  and Mechanisms, IFToMM, Sevilla, Spain}, 1987.

\bibitem[Parikh et~al.(2014)Parikh, Boyd, et~al.]{parikh2014proximal}
Neal Parikh, Stephen Boyd, et~al.
\newblock Proximal algorithms.
\newblock \emph{Foundations and trends{\textregistered} in Optimization},
  1\penalty0 (3):\penalty0 127--239, 2014.

\bibitem[Popov et~al.(1978)Popov, Vereshchagin, and
  Zenkevi{\v{c}}]{popov1978manipuljacionnyje}
Je~P Popov, Anatolij~Fedorovi{\v{c}} Vereshchagin, and
  Stanislav~Leonidovi{\v{c}} Zenkevi{\v{c}}.
\newblock \emph{Manipuljacionnyje roboty: Dinamika i algoritmy}.
\newblock Nauka, 1978.

\bibitem[Rawlings et~al.(2017)Rawlings, Mayne, and Diehl]{rawlings2017model}
James~Blake Rawlings, David~Q Mayne, and Moritz Diehl.
\newblock \emph{Model predictive control: theory, computation, and design},
  volume~2.
\newblock Nob Hill Publishing Madison, 2017.

\bibitem[Sathya and Carpentier(2024)]{sathya2024constrained}
Ajay Sathya and Justin Carpentier.
\newblock Constrained articulated body dynamics algorithms.
\newblock \emph{Conditionally Accepted to IEEE Transactions on Robotics}, 2024.

\bibitem[Sathya et~al.(2022)Sathya, Astudillo, Gillis, Decr{\'e}, Pipeleers,
  and Swevers]{sathya2022tasho}
Ajay~Suresha Sathya, Alejandro Astudillo, Joris Gillis, Wilm Decr{\'e}, Goele
  Pipeleers, and Jan Swevers.
\newblock Tasho: A python toolbox for rapid prototyping and deployment of
  optimal control problem-based complex robot motion skills.
\newblock In \emph{2022 IEEE/RSJ International Conference on Intelligent Robots
  and Systems (IROS)}, pages 9700--9707. IEEE, 2022.

\bibitem[Sathya et~al.(2024{\natexlab{a}})Sathya, Bruyninckx, Decré, and
  Pipeleers]{sathya_constrained_dynamics}
Ajay~Suresha Sathya, Herman Bruyninckx, Wilm Decré, and Goele Pipeleers.
\newblock Efficient constrained dynamics algorithms based on an equivalent lqr
  formulation using gauss' principle of least constraint.
\newblock \emph{IEEE Transactions on Robotics}, 40, 2024{\natexlab{a}}.

\bibitem[Sathya et~al.(2024{\natexlab{b}})Sathya, Decre, and
  Swevers]{sathya_low_order_delassus}
Ajay~Suresha Sathya, Wilm Decre, and Jan Swevers.
\newblock Pv-osimr: A lowest order complexity algorithm for computing the
  delassus matrix.
\newblock \emph{arXiv preprint arXiv:2310.03676}, 2024{\natexlab{b}}.
\newblock In revision for IEEE RA-L.

\bibitem[Sherman and Morrison(1950)]{sherman1950adjustment}
Jack Sherman and Winifred~J Morrison.
\newblock Adjustment of an inverse matrix corresponding to a change in one
  element of a given matrix.
\newblock \emph{The Annals of Mathematical Statistics}, 21\penalty0
  (1):\penalty0 124--127, 1950.

\bibitem[Suh et~al.(2022{\natexlab{a}})Suh, Simchowitz, Zhang, and
  Tedrake]{suh2022differentiable}
Hyung~Ju Suh, Max Simchowitz, Kaiqing Zhang, and Russ Tedrake.
\newblock Do differentiable simulators give better policy gradients?
\newblock In \emph{International Conference on Machine Learning}, pages
  20668--20696. PMLR, 2022{\natexlab{a}}.

\bibitem[Suh et~al.(2022{\natexlab{b}})Suh, Pang, and Tedrake]{suh2022bundled}
Hyung Ju~Terry Suh, Tao Pang, and Russ Tedrake.
\newblock Bundled gradients through contact via randomized smoothing.
\newblock \emph{IEEE Robotics and Automation Letters}, 7\penalty0 (2):\penalty0
  4000--4007, 2022{\natexlab{b}}.

\bibitem[Sutton and Barto(2018)]{sutton2018reinforcement}
Richard~S Sutton and Andrew~G Barto.
\newblock \emph{Reinforcement learning: An introduction}.
\newblock MIT press, 2018.

\bibitem[Tassa et~al.(2014)Tassa, Mansard, and Todorov]{tassa2014control}
Yuval Tassa, Nicolas Mansard, and Emo Todorov.
\newblock Control-limited differential dynamic programming.
\newblock In \emph{2014 IEEE International Conference on Robotics and
  Automation (ICRA)}, pages 1168--1175. IEEE, 2014.

\bibitem[Tedrake and the Drake Development~Team(2019)]{drake}
Russ Tedrake and the Drake Development~Team.
\newblock Drake: Model-based design and verification for robotics, 2019.
\newblock URL \url{https://drake.mit.edu}.

\bibitem[Todorov(2014)]{todorov2014convex}
Emanuel Todorov.
\newblock Convex and analytically-invertible dynamics with contacts and
  constraints: Theory and implementation in mujoco.
\newblock In \emph{Proc. IEEE Int. Conf. Robot. Autom.}, pages 6054--6061.
  IEEE, 2014.

\bibitem[Todorov et~al.(2012)Todorov, Erez, and Tassa]{todorov2012mujoco}
Emanuel Todorov, Tom Erez, and Yuval Tassa.
\newblock Mujoco: A physics engine for model-based control.
\newblock In \emph{Proc. IEEE/RSJ Int. Conf. Int. Robots. Syst.}, pages
  5026--5033. IEEE, 2012.

\bibitem[Toussaint et~al.()Toussaint, Allen, Smith, and
  Tenenbaum]{toussaintdifferentiable}
Marc Toussaint, Kelsey~R Allen, Kevin~A Smith, and Joshua~B Tenenbaum.
\newblock Differentiable physics and stable modes for tool-use and manipulation
  planning.

\bibitem[Touvron et~al.(2023)Touvron, Martin, Stone, Albert, Almahairi, Babaei,
  Bashlykov, Batra, Bhargava, Bhosale, et~al.]{touvron2023llama}
Hugo Touvron, Louis Martin, Kevin Stone, Peter Albert, Amjad Almahairi, Yasmine
  Babaei, Nikolay Bashlykov, Soumya Batra, Prajjwal Bhargava, Shruti Bhosale,
  et~al.
\newblock Llama 2: Open foundation and fine-tuned chat models.
\newblock \emph{arXiv preprint arXiv:2307.09288}, 2023.

\bibitem[Udwadia and Kalaba(1996)]{udwadia1996}
Firdaus~E Udwadia and Robert~E Kalaba.
\newblock \emph{Analytical dynamics : a new approach}.
\newblock Cambridge University press, Cambridge, 1996.
\newblock ISBN 0-521-48217-8.

\bibitem[Vereshchagin(1989)]{vereshchagin1989modeling}
Anatolii~Fedorovich Vereshchagin.
\newblock Modeling and control of motion of manipulational robots.
\newblock \emph{Soviet Journal of Computer and Systems Sciences}, 27\penalty0
  (5):\penalty0 29--38, 1989.

\bibitem[Wensing et~al.(2012)Wensing, Featherstone, and
  Orin]{wensing2012reduced}
Patrick Wensing, Roy Featherstone, and David~E Orin.
\newblock A reduced-order recursive algorithm for the computation of the
  operational-space inertia matrix.
\newblock In \emph{Proc. IEEE Int. Conf. Robot. Autom.}, pages 4911--4917.
  IEEE, 2012.

\end{thebibliography}

\end{document}